\documentclass[conference]{IEEEtran}
\usepackage{amsmath,amssymb,amsfonts}
\usepackage{algorithm} 
\usepackage{algpseudocode}
\usepackage{graphicx}
\usepackage{float}
\usepackage{textcomp}
\usepackage{booktabs}
\usepackage{xcolor}
\usepackage{array}
\usepackage{natbib}
\setcitestyle{authoryear,open={(},close={)}}

\usepackage{hyperref}

\begin{document}

\title{Motion Primitives based Path Planning with Rapidly-exploring Random Tree}

\author{\IEEEauthorblockN{Abhishek Paudel}
\IEEEauthorblockA{
Department of Computer Science \\
George Mason University \\
Fairfax, Virginia, USA}
}

\maketitle

\begin{abstract}
We present an approach that generates kinodynamically feasible paths for robots using Rapidly-exploring Random Tree (RRT). We leverage motion primitives as a way to capture the dynamics of the robot and use these motion primitives to build branches of the tree with RRT. Since every branch is built using the robot's motion primitives that doesn't lead to collision with obstacles, the resulting path is guaranteed to satisfy the robot's kinodynamic constraints and thus be feasible for navigation without any post-processing on the generated trajectory. We demonstrate the effectiveness of our approach in simulated 2D environments using simple robot models with a variety of motion primitives.

\end{abstract}


\section{Introduction} \label{sec:introduction}
Path planning is one of the fundamental problems in robotics. It deals with finding a feasible path from a starting point to a goal point in a configuration space while avoiding collisions with obstacles. Path planning approaches can generally be classified into two categories: \emph{exact} and \emph{sampling-based} methods \citep{lavalle2006planning}. While exact methods look for solution directly in continuous configuration space, sampling-based methods tend to only look at sampled states of the configuration space and generally construct a graph of these samples in an attempt to find a solution. Although exact methods are said to be complete, they are PSPACE-hard \citep{reif1979complexity} and hence difficult to solve even for low dimensional configuration space. Therefore, sampling-based methods have emerged as popular approaches for path planning because of their practical applicability in various scenarios. Rapidly-exploring Random Tree (RRT) \citep{lavalle1998rapidly} is one of the widely used sampling-based methods for path planning in which a search tree is built by repeatedly sampling random points and adding branches towards them from the nearest node in the tree. RRT builds the tree with the starting point as the root node and  the tree spans the configuration space. The path from start to goal is obtained by following the branches from the goal point to the root node i.e., starting point.

In general, the branches in RRT are straight lines paths connecting two nodes in the tree. Therefore, the final path from starting point to goal point obtained using RRT is a set of straight lines with sharp or angular turns at the nodes. However, depending on the dynamics of a mobile robot, thus obtained path might be infeasible for the robot to navigate since the robot may not be able to take sharp turns. A common approach in such scenarios is to smooth the obtained path so that the smoothed path is feasible for the robot to navigate through \citep{yang2010analytical, ravankar2018path}. However, this adds an extra post-processing step on top of existing path planners and depending on the robot, path smoothing may not be a feasible option to handle all kinds of vehicle dynamics. Therefore, it is important that existing path planning approaches be able to handle vehicle dynamics so that the obtained path is always a feasible trajectory that the robot can navigate through.

In this paper, we propose to use motion primitives as a natural way to incorporate vehicle dynamics and use collision-free motion primitives to build branches of the tree with RRT. Since the obtained path from start to goal is a chain of collision-free motion primitives, thus obtained path satisfy the kinodynamic constraints of the robot and is always feasible for navigation.
\begin{figure}[t]
    \centering
    \includegraphics[width=8cm]{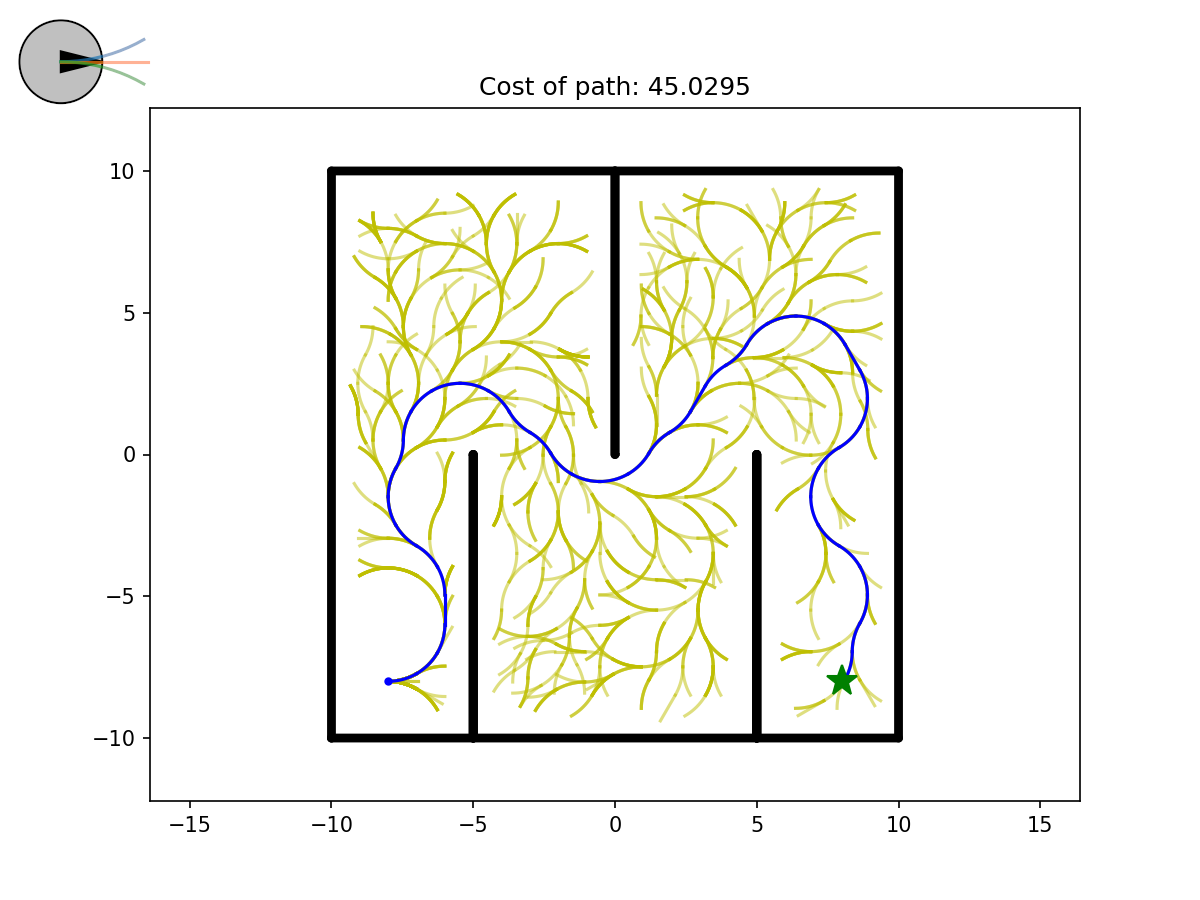}
    \caption{An example of path obtained using our motion primitives based RRT. Motion primitives shown on the top left consist of three actions: move forward, turn 30 degrees towards left with a radius of curvature of 2 units, and turn 30 degrees towards right with a radius of curvature of 2 units.}
    \label{fig:dubins_pi6_2p0}
\end{figure}

The rest of the paper is organized as follows. Section \ref{sec:related_work} discusses related work in this domain. Section \ref{sec:motion_primitives} discusses motion primitives in the context of our approach. Section \ref{sec:rrt_with_motion_primitives} discusses our approach on motion primitives based RRT. Section \ref{sec:experiments} discusses our experiments and their results. Finally, Section \ref{sec:conclusion} includes concluding remarks and potential future work extending from this work.

\section{Related Work} \label{sec:related_work}
Incorporating vehicle dynamics for path planning has been studied widely across the robotics literature. Many sampling-based methods for path planning result in jagged paths for traversal. Many approaches try to tackle this by applying smoothing techniques on the paths generated by path planners so as to make them feasible for non-holonomic systems \citep{garrido2009smooth, elbanhawi2015continuous}. Path smoothing specific to robot navigation is also widely studied in robotics of which \citet{ravankar2018path} present a detailed survey. However, path smoothing is a post-processing step on top of path planners, and there are approaches that try to avoid this by modeling smooth paths during path planning. A notable example of this approach is presented in \citet{lau2009kinodynamic} in which splines based on vehicle dynamics are used to join two way-points during path planning so as to obtain feasible paths. Such approach which tries to combine the search for a collision-free path by taking into account the dynamics of the robot so as find a feasible trajectory is also known in robotics literature as \emph{kinodynamic planning} \citep{donald1993kinodynamic}.

Another widely used model to plan smooth paths is the Dubin's curve \citep{dubins1957oncurves} which is used to model the trajectory of a car (eponymously known as the Dubin's car) that can either go forward, steer all the way to left or steer all the way to right. \citet{vzivojevic2019path} propose to use the Dubin's car model for path planning using RRT and demonstrate that their approach generates navigable path for Dubin's car like robots. Specific to RRT, \citet{palmieri2014efficient} propose a modified extend function for RRT to generate smooth path between any two states and thus generate a path that is feasible for navigation. Other approaches that discuss RRT-based path planners to account for robot's dynamics include \citet{hu2020efficient} and \citet{kuwata2008motion}. There are also approaches that consider motion primitives in some way for path planning using sampling-based methods \citep{sakcak2018using, sakcak2019sampling}.

Our approach focuses specifically on using motion primitives as a way to incorporate arbitrary vehicle dynamics, and use these motion primitives extend the tree in RRT so as to find a path that is feasible for the given vehicle dynamics.

\section{Motion Primitives} \label{sec:motion_primitives}
\begin{figure}[ht]
    \centering
    \includegraphics[width=8cm]{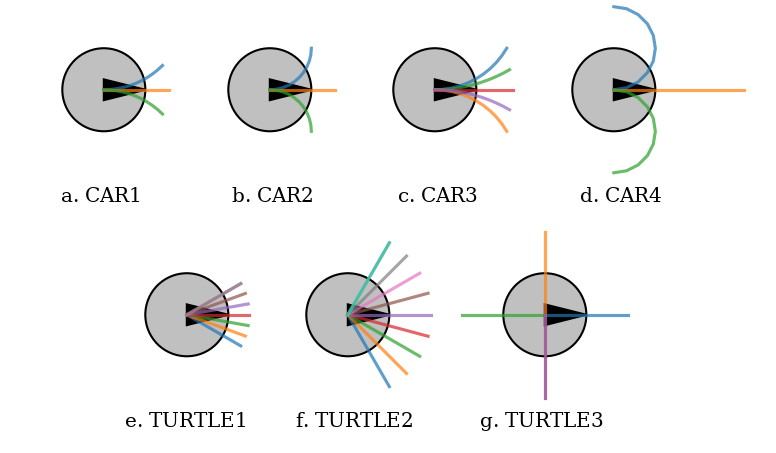}
    \caption{A demonstration of car-like (top) and turtlebot-like (bottom) motion primitives designed for our experiments.}
    \label{fig:motionprimitives}
\end{figure}

Motion primitives are a set of discrete pre-computed motions that the robot can take from a given state. Motion primitives are used to capture the dynamics of a robot in a simple and elegant way. They define the immediate possible trajectories that the robot can take based on available control inputs. Since each motion primitive is a dynamically feasible trajectory for the robot, any path computed for combining a sequence of motion primitives is also dynamically feasible for traversal. We design a variety of car-like and turtlebot-like motion primitives as shown in Fig \ref{fig:motionprimitives}, and use these motion primitives for path planning using our approach. Car-like motion primitives are representative of dynamics in which the robot cannot turn in place and has to take a curved path to make a turn. Turtlebot-like motion primitives are representative of dynamics in which the robot can only turn in place and move in straight lines. Each of these motion primitives are described below.

\begin{itemize}
    \item \textbf{CAR1}: The robot can either go forward without turning or make turns of 45 degrees in left and right directions as it goes forward with a radius of curvature of 1 units.
    \item \textbf{CAR2}: The robot can either go forward without turning or make turns of 90 degrees in left and right directions as it goes forward with a radius of curvature of 0.5 units.
    \item \textbf{CAR3}: The robot can either go forward without turning or make turns of 30 and 60 degrees in left and right directions as it goes forward with a radius of curvature of 1.8 units and 1 units respectively. 
    \item \textbf{CAR4}: The robot can either go forward without turning or make turns of 180 degrees in left and right directions as it goes forward with a radius of curvature of 0.5 units.
    \item \textbf{TURTLE1}: The robot can move forward after making in place turns of angles of 10, 20 and 30 degrees in left and right directions or move forward without making any in place turns.
    \item \textbf{TURTLE2}: The robot can move forward after making in place turns of angles of 15, 30, 45 and 60 degrees in left and right directions or move forward without making any in place turns.
    \item \textbf{TURTLE3}: The robot can move forward after making in place turns of angles of 90 or 180 degrees in left direction and 90 degrees in right direction or move forward without making any in place turns.
\end{itemize}

\section{RRT with Motion Primitives} \label{sec:rrt_with_motion_primitives}
In the basic form of RRT, the search tree is incrementally built from the starting point and every branch added to the tree is a straight line that connects two nodes if there is no collision with obstacles. This results in path that has sharp or angular turns making it infeasible for robots that cannot take sharp turns. To alleviate this, we use the robot's motion primitives to build the search tree. Following the standard RRT approach, we sample a random point and pick the node in the tree that is nearest to the sampled point. To extend the tree from this nearest node, we look at the available motion primitives in that node that do not lead to collisions with obstacles. Using these motion primitives, we transform the robot pose at the nearest node by applying each of these motion primitives to get a new set of poses and pick the pose that moves the robot closest to the sampled point. A feasible path from start to goal can then be generated by following the branches in the tree from goal to start.

\begin{algorithm}
\caption{Motion-Primitives-RRT($q_0, prim, obs, K$)}
\label{alg:motion_primitives_rrt}
    \begin{algorithmic}
        \State $\mathcal{G}.$init$(q_0)$
        \For{$i$ = 1 to $K$}
            \State $q_{rand} \leftarrow $ RandomPoint$(\mathcal{C})$
            \State $q_{near} \leftarrow $ NearestNode$(q_{rand},\mathcal{G})$
            \State $feasible\_prim \leftarrow $ CollisionCheck$(q_{near}, prim, obs)$
            \State $poses \leftarrow $ Transform$(q_{near}, feasible\_prim)$
            \State $q_{new} \leftarrow $ ClosestPose$(poses, q_{rand})$
            \State Extend$(\mathcal{G},q_{near},q_{new})$
        \EndFor
        \State \Return $\mathcal{G}$
    \end{algorithmic}
\end{algorithm}

\section{Results} \label{sec:experiments}
\begin{figure}[H]
    \centering
    \includegraphics[width=8.5cm]{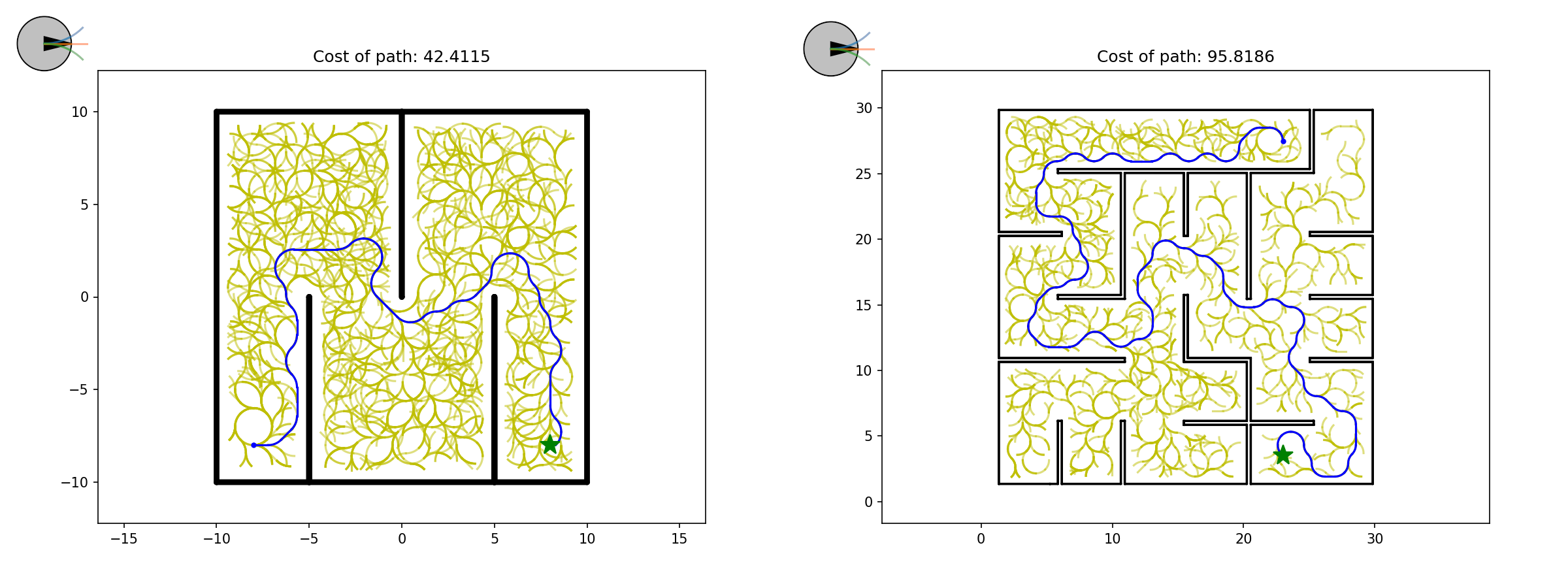}
    \caption{Path obtained for robot with CAR1 motion primitives}
    \label{fig:dubins_pi4_1p0}
\end{figure}

\begin{figure}[H]
    \centering
    \includegraphics[width=8.5cm]{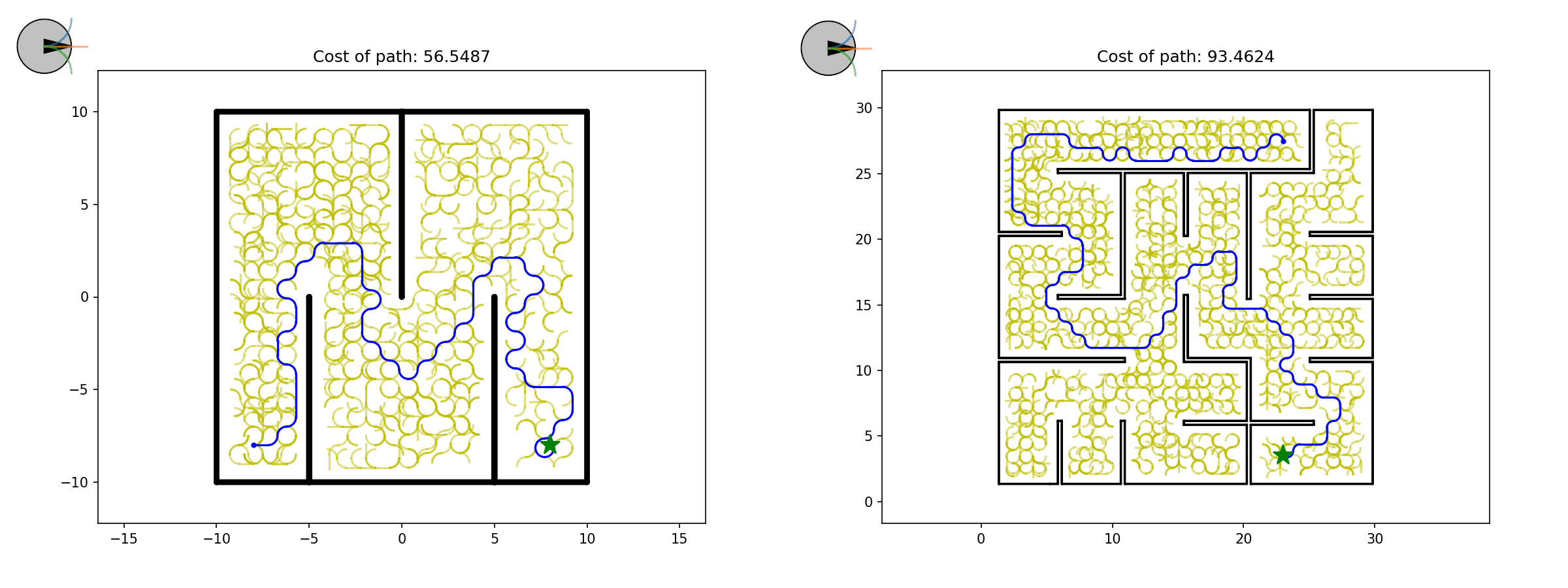}
    \caption{Path obtained for robot with CAR2 motion primitives}
    \label{fig:dubins_pi2_0p5}
\end{figure}

\begin{figure}[H]
    \centering
    \includegraphics[width=8.5cm]{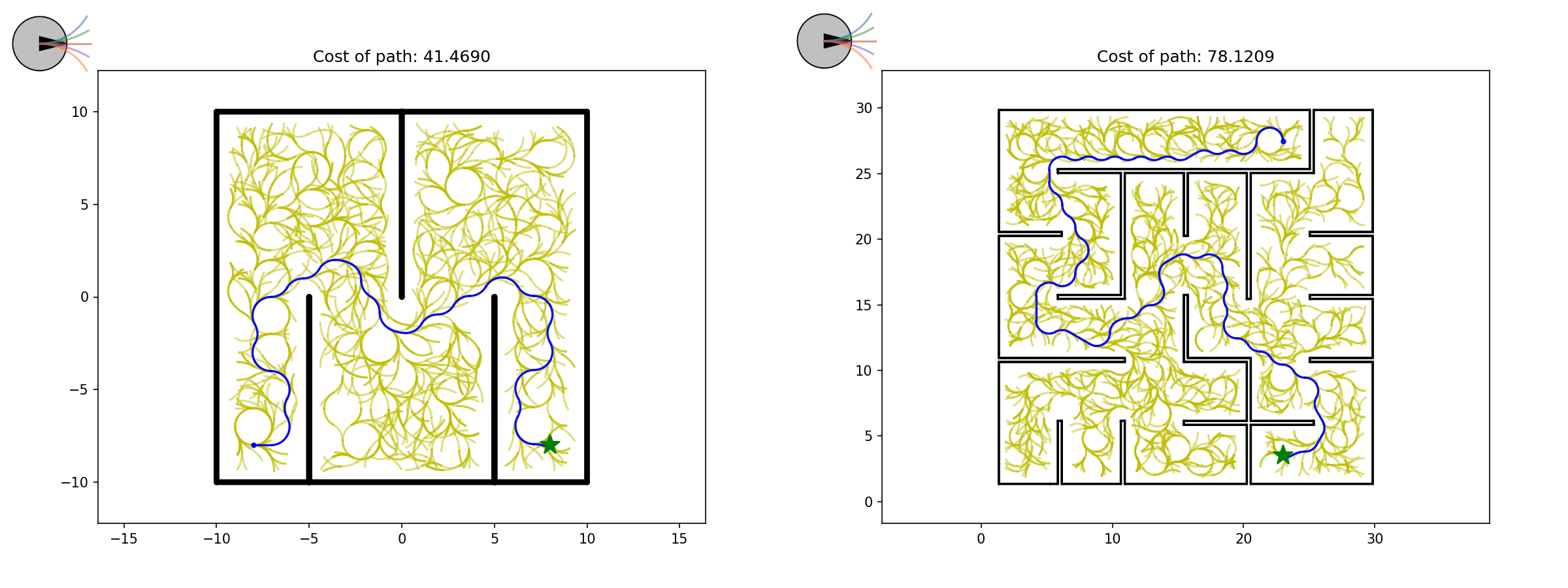}
    \caption{Path obtained for robot with CAR3 motion primitives}
    \label{fig:dubins_many}
\end{figure}

\begin{figure}[H]
    \centering
    \includegraphics[width=8.5cm]{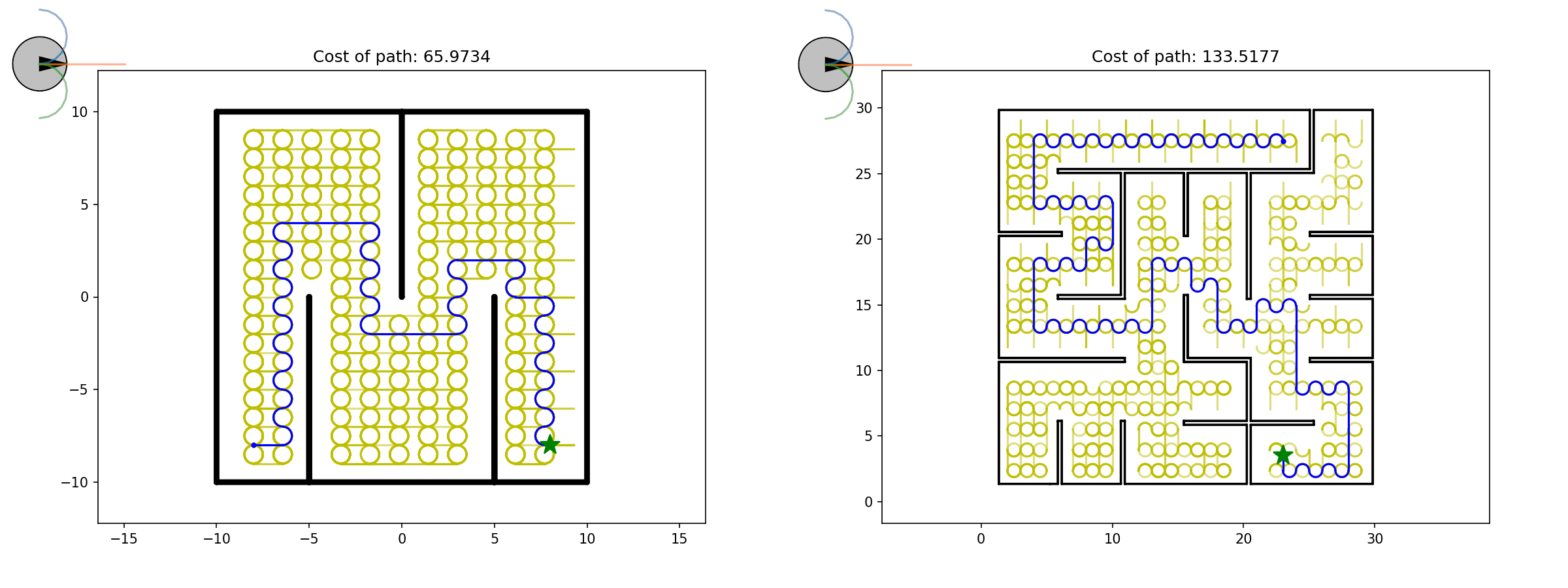}
    \caption{Path obtained for robot with CAR4 motion primitives}
    \label{fig:dubins_pi_0p5}
\end{figure}

\begin{figure}[H]
    \centering
    \includegraphics[width=8.5cm]{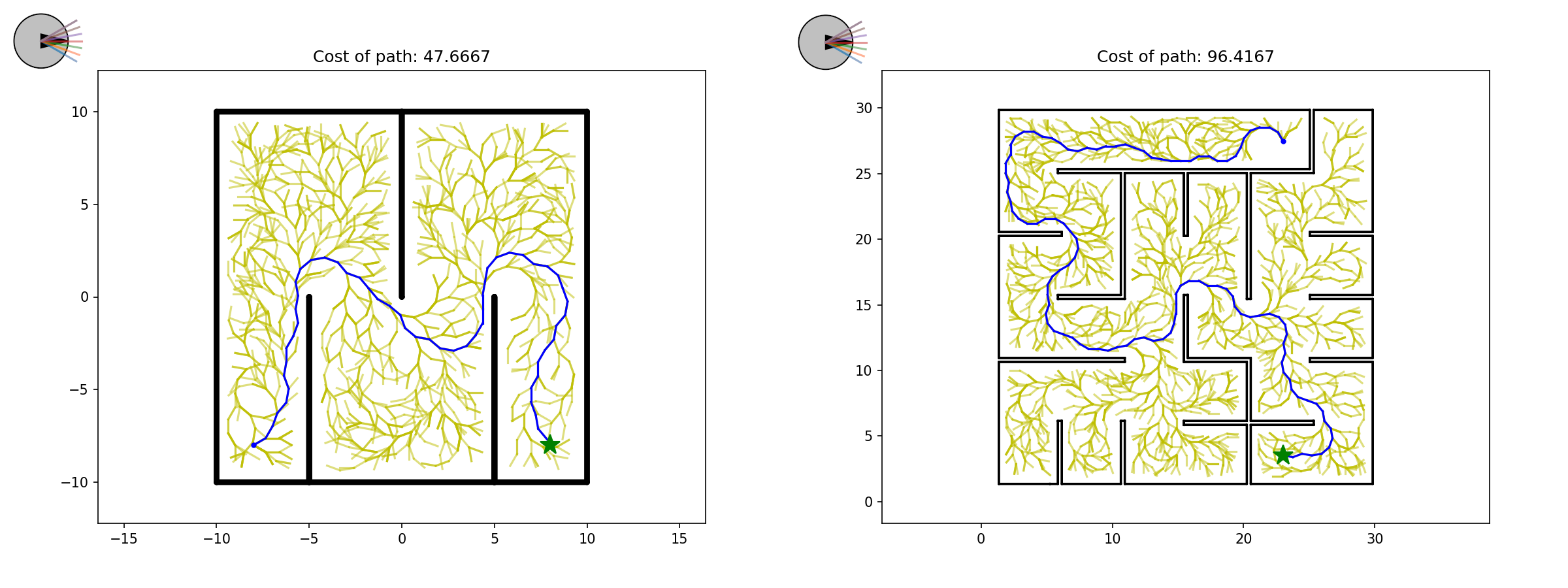}
    \caption{Path obtained for robot with TURTLE1 motion primitives}
    \label{fig:turtlebot_pi3_0p75_3}
\end{figure}

\begin{figure}[H]
    \centering
    \includegraphics[width=8.5cm]{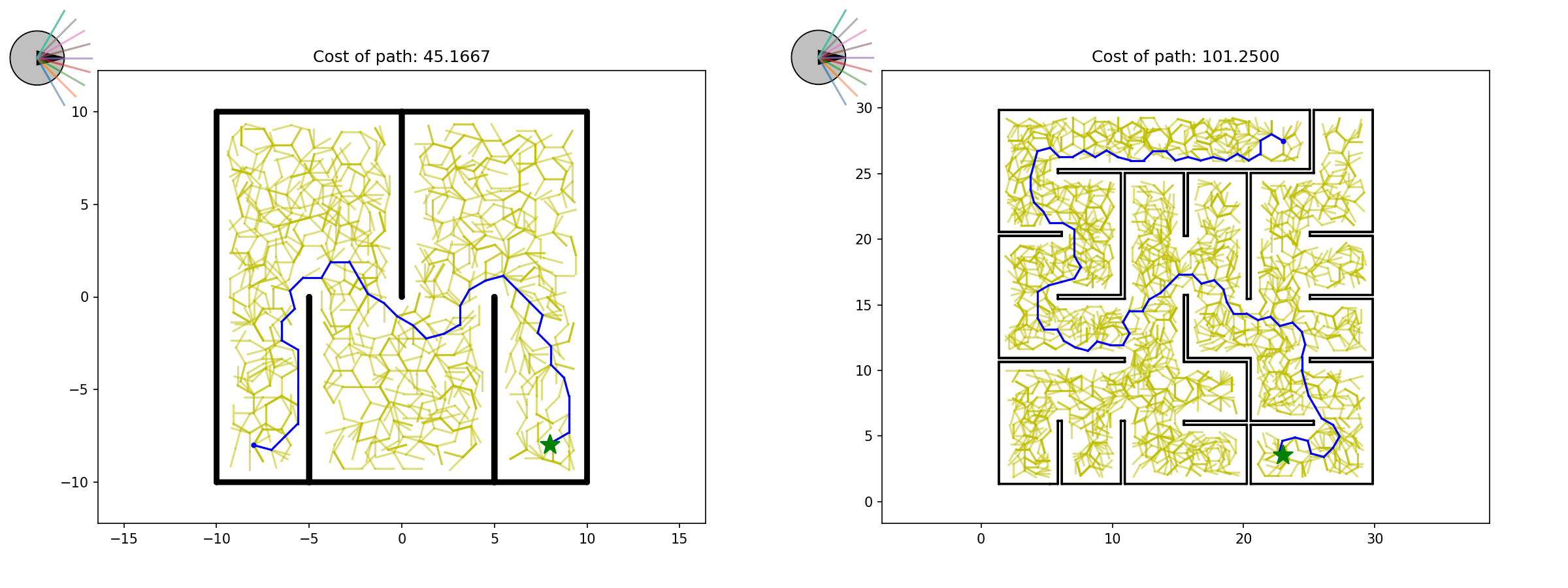}
    \caption{Path obtained for robot with TURTLE2 motion primitives}
    \label{fig:turtlebot_pi1p5_1p0_4}
\end{figure}

\begin{figure}[H]
    \centering
    \includegraphics[width=8.5cm]{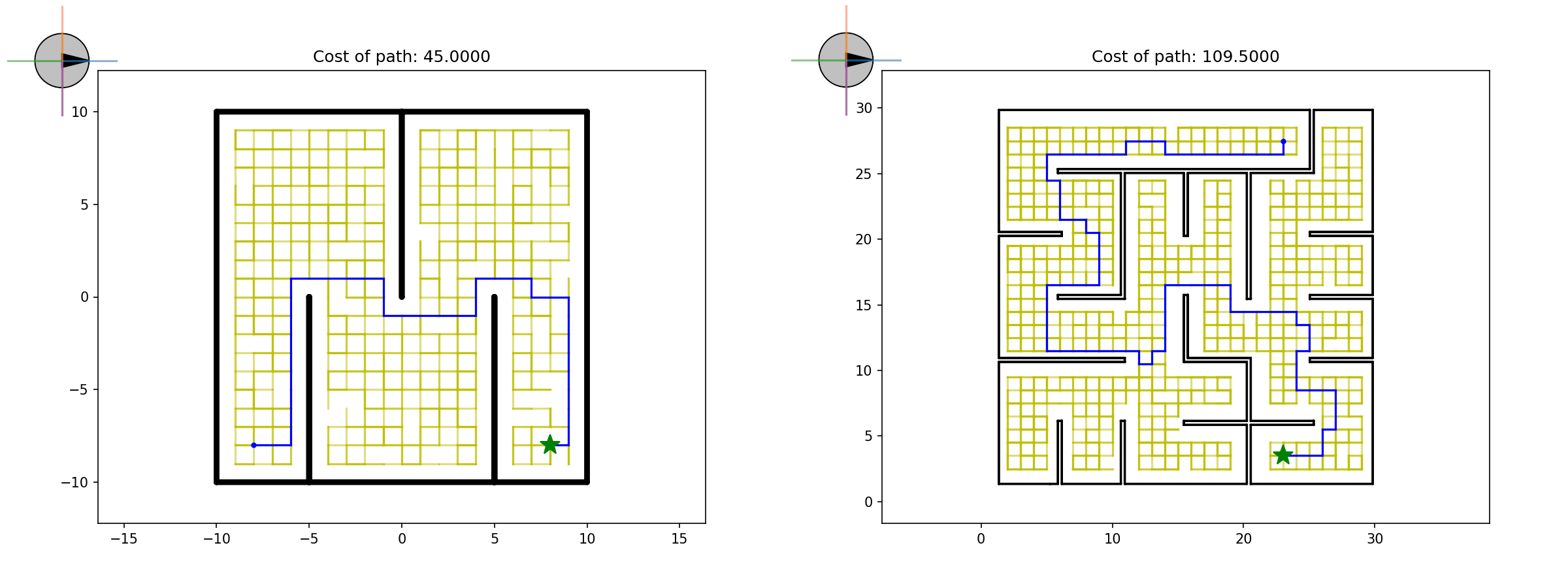}
    \caption{Path obtained for robot with TURTLE3 motion primitives}
    \label{fig:perp_turtlebot_1p0}
\end{figure}

We evaluate our approach in small and large 2D maze environments to demonstrate its effectiveness. The simulation experiments are done with a circular robot model of radius 0.5 units. To avoid path that leads to collision due to robot's size, path planning is done by inflating the obstacles by the robot's radius and assuming robot to be a point. Figure \ref{fig:dubins_pi4_1p0} - \ref{fig:perp_turtlebot_1p0} show the path obtained using our approach in the two environments for different motion primitives. Animation of simulated robots navigating in the maze environments on the obtained paths can be viewed on this URL\footnote{\url{https://imgur.com/a/pH9Xx6m}}.


\section{Conclusion and Future Work} \label{sec:conclusion}
We presented motion primitives based path planning approach using RRT that handles a wide variety of mobile robot dynamics. Robot dynamics are captured in the motion primitives as a set of discrete pre-computed motions that are feasible for the robot in a given state, and a modified RRT algorithm is used to generate a kinodynamically feasible path from start to goal. Our approach is general enough to handle a wide variety of motion primitives and also dynamic motion primitives that are a function of robot state. However, since our approach is an extension of RRT, the obtained path is not guaranteed to be asymptotically optimal. Extending our approach to RRT* could be one way to get optimal path, but this would require rewiring of branches under the constraints of available motion primitives. This is a non-trivial task and can be explored in future work.

\bibliography{references.bib}

\end{document}